# PIKFNO: An Interpretable Neural Operator Based on Physics-Informed Kernel Function

Yuan Guo[a], Hanshu Chen[a], Zhuojia Fu[a,b,*]

[a] *College of Mechanics and Engineering Science, Hohai University, Nanjing 211100, China*
[b] *Key Laboratory of Ministry of Education for Coastal Disaster and Protection, Hohai University, Nanjing 210098, China*

**Abstract:** This work proposes a new interpretable neural operator framework, termed the Physics-Informed Kernel Function Neural Operator (PIKFNO), which explicitly incorporates physics-informed kernel functions derived from governing equations into the neural operator architecture. Unlike traditional neural operators such as DeepONet, which rely on deep networks to implicitly learn basis functions, PIKFNO constrains the trunk network through physics-informed kernel functions, thereby aligning its operator structure with the kernel expansions used in meshless collocation methods. Two construction strategies are introduced: one learns kernel functions directly from data, where the learned kernel can be regarded as a nonsingular fundamental solution, while the other builds them through transformations of analytical fundamental solutions. Numerical experiments demonstrate that PIKFNO achieves high predictive accuracy with substantially improved interpretability and superior generalization under limited training data. The proposed framework offers a new pathway for developing efficient, physically consistent, and interpretable neural operators.

## 1 Introduction

In recent years, the rapid advancement of machine learning has profoundly influenced the study of partial differential equations (PDEs), leading to the emergence of neural operators as a powerful framework. Unlike traditional numerical schemes, such as finite element methods[1], finite difference methods[2] or meshless methods[3], that rely on discretized grids or predefined basis functions, neural operators aim to directly learn mappings between infinite-dimensional function spaces, thereby enabling end-to-end approximation of PDE solution operators[4]. These methods have demonstrated remarkable advantages in multiscale modeling, complex physical system simulation, and real-time prediction[5]. Representative frameworks, including DeepONet[6] and Fourier Neural Operator (FNO)[7], have attracted extensive attention in fields such as mechanics[8–10], acoustics [11],electromagnetics[12], and materials science[13].

Despite their success in accuracy and efficiency, the limited interpretability of neural operators remains a major obstacle to their broader adoption in scientific and engineering applications. On the one hand, mainstream neural operators rely on implicit representations learned by deep networks, making it difficult to establish clear correspondences with operator structures in traditional numerical analysis. On the other hand, the absence of explicit physical constraints often leads to large data requirements, poor extrapolation capability, violations of physical laws, or sensitivity to noise. Consequently, developing neural operators that simultaneously achieve interpretability, physical consistency, and computational efficiency has become a central research challenge.

[*] Corresponding author
E-mail address: paul212063@hhu.edu.cn (Z. Fu).

To enhance physical consistency, various physics-informed learning frameworks have been proposed, such as Physics-Informed Neural Networks (PINNs)[14], physics-informed DeepONet[15], and structured networks incorporating conservation laws or symmetries[16–18]. Although these approaches alleviate the "black-box" issue to some extent, they still suffer from two major limitations: (i) physical constraints are typically imposed through loss terms, functioning as soft constraints that do not guarantee physical consistency at the operator level; and (ii) the internal representations remain implicit, making it difficult to provide mathematically interpretable operator structures.

Meanwhile, kernel methods have long played a central role in numerical analysis and machine learning[19]. Approaches such as radial basis function (RBF) methods, meshless collocation methods, and Mercer-type kernel expansions possess clear mathematical formulations and strong interpretability[20]. Motivated by these observations, we propose a new interpretable neural operator framework— Physics-Informed Kernel Function Neural Operator (PIKFNO). The core ideas of PIKFNO are summarized as follows:

(1) Constructing operator structures using physics-informed kernel functions. These kernel functions explicitly encode the structural characteristics of the governing PDE, ensuring that the resulting operator remains consistent with kernel integral formulations and naturally incorporates physical constraints.

(2) Achieving flexible expressiveness through learnable kernel functions. By introducing learnable parameters while preserving the interpretability of kernel methods, the model combines the mathematical transparency of classical kernel approaches with the expressive power of neural networks.

(3) Embedding physical priors at the operator level to enhance physical consistency. By explicitly incorporating physical laws into the operator construction, the proposed framework achieves superior physical fidelity and generalization performance compared with traditional neural operator methods.

# 2 Methodology

## 2.1 Operator Learning

In scientific computing and engineering applications, many fundamental problems can be abstracted as mappings from functions to functions. In particular, a partial differential equation (PDE) inherently defines an operator that maps input functions—such as coefficient fields, source terms, or boundary conditions—to output functions, typically the solution fields. For example, in elliptic PDEs, the solution process can be viewed as an operator that maps boundary conditions to interior solutions;

for time-dependent problems, it corresponds to an evolution operator that maps an initial state to future states. Therefore, learning the solution process of a PDE is, in essence, equivalent to learning an unknown and potentially highly nonlinear operator.

Traditional neural network approaches typically approximate mappings between discretized data, and their performance often depends on the resolution of the input and output grids. When the mesh density changes, the model usually requires retraining or explicit interpolation, making it difficult to maintain consistency across different discretization levels. In contrast, the goal of operator learning is to learn a mapping that acts between function spaces, with discretization invariance as a key characteristic: the learned operator remains consistent even when the input and output functions are represented on different grids, sampling points, or resolutions [7]. However, not all data-driven models qualify as neural operators. To be considered a genuine neural operator, a model must not only exhibit discretization invariance but also satisfy the theoretical requirement of universal approximation of operators.

## 2.2 DeepONet

DeepONet is one of the earliest neural operator frameworks that successfully implemented operator learning. Its core idea is rooted in the universal approximation theorem for operators on Banach spaces. The architecture of DeepONet consists of two components: a Branch Net and a Trunk Net. The Branch Net takes as input the discrete samples of the function $u(x)$ and extracts its global features, while the Trunk Net receives the target location $y$ and learns the spatial representation of a set of basis functions. The final output is obtained by combining the two through an inner-product structure:

$$G(u)(y) \approx \sum_{k=1}^{p} b_k(u) t_k(y) \tag{1}$$

where $b_k(u)$ is generated by the Branch Net and $t_k(y)$ is produced by the Trunk Net. This formulation essentially corresponds to a kernel expansion or Mercer expansion of an operator, providing a clear mathematical interpretation: DeepONet constructs operator approximations through learnable basis functions and learnable coefficients.

Unlike traditional neural networks, DeepONet is inherently designed to satisfy two key properties of operator learning: discretization invariance and universal approximation of operators. Since the Branch Net takes function samples rather than fixed-resolution grid data, the input resolution can vary; as long as the sampling points sufficiently represent the input function, the learned operator remains consistent across different discretization levels. Moreover, because the architecture directly mirrors the basis-function expansion used in operator theory, DeepONet is theoretically capable of approximating any continuous operator. For PDE problems, where the governing equation defines a mapping from boundary conditions, initial conditions, or coefficient fields to solution fields, DeepONet can directly learn this mapping, giving it a natural advantage in PDE solving.

Nevertheless, DeepONet also has certain limitations. For example, the basis functions $t_k(y)$ learned by the Trunk Net lack explicit physical meaning, and the coefficients $b_k(u)$ produced by the Branch Net are difficult to interpret in terms of PDE-related physical quantities. Thus, despite its strong expressive power in operator approximation, the internal structure of DeepONet remains largely a "black box," with limited interpretability.

### 2.3 Meshless Collocation Method and Physics-Informed Kernel Function

The "coefficient–basis" expansion adopted by DeepONet bears a natural structural resemblance to meshless collocation methods in traditional numerical analysis. In meshless methods, the solution of a partial differential equation (PDE) is typically represented as a linear combination of kernel functions, where the kernels are determined by the operator structure of the governing equation. For example, in the Method of Fundamental Solutions (MFS)[21] or boundary value problems, the basis functions are chosen as the fundamental solution $\varphi(r)$ of the PDE. Given a set of source points $X_s^i$, the solution corresponding to any boundary condition $u_b$ can be expressed as

$$G(u_b)(X) \approx \sum_{i=1}^{n} \omega_i(u_b)\varphi(\| X - X_s^i \|) \tag{2}$$

where the coefficients $\omega_i$ are uniquely determined by the boundary condition. This representation is essentially a discretization of a kernel integral operator, in which the kernel $\varphi(r)$ directly reflects the characteristics of the PDE operator and can be regarded as a physics-informed kernel function. where the kernels are determined by the operator structure of the governing equation. These kernels inherently embed the physical information of the governing equations. Some governing equations and their corresponding physics-informed kernel function can be found in[22]. For instance, for the two-dimensional Laplace equation, the fundamental solution is $\varphi(r) = -\frac{1}{2\pi}\ln(r)$, which corresponds to the Green's function of the Laplace operator. This reveals a deep connection between meshless methods and operator learning: both approximate the mapping from input functions to output functions through expansions in basis functions.

Despite the high accuracy and mesh-free nature of MFS, its practical application faces several challenges, among which the selection of source point locations is particularly critical. Improper placement of source points may lead to a severely ill-conditioned coefficient matrix, causing large variations in the coefficients $\omega_i$ and compromising numerical stability. To mitigate this issue, a common strategy is to place the source points directly on the boundary, leading to the Singular Boundary Method (SBM)[23]. However, because the fundamental solution is singular on the boundary—i.e., $\varphi(r) \to \infty$ as $r \to 0$—SBM requires additional techniques to handle the singularity, such as singularity subtraction or local coordinate transformations.

### 2.4 Physics-Informed Kernel Function Neural Operator

In meshless collocation methods, physics-informed kernel functions are typically derived directly from the governing equations and therefore naturally encode the physical properties of the PDE. However, for many complex governing equations, such kernel functions are either difficult to derive, mathematically intractable, or may not exist in closed form. Meanwhile, in practical engineering applications, large quantities of measurement data or numerical solution samples are often available. This raises an important question: Can physics-informed kernel functions be learned directly from data? Motivated by this idea, we propose the Physics-Informed Kernel Function Neural Operator (PIKFNO), which aims to explicitly incorporate physical structures into the neural operator framework, thereby enhancing interpretability and reducing the amount of training data required.

The design of PIKFNO is directly inspired by the structural correspondence between meshless methods and DeepONet, as illustrated in Figure1. In MFS, the kernel function $\varphi(r)$ is determined by the PDE operator, while the coefficients $\omega_i(u_b)$ are uniquely determined by the boundary condition. In contrast, DeepONet learns the basis functions $t_i(X)$ through the trunk network and generates the coefficients $b_i(u_b)$ through the branch network. Although the two approaches share a highly similar functional form, MFS provides explicit physical meaning, whereas DeepONet offers strong data-driven expressiveness. Therefore, combining "physics-informed kernel functions determined by the PDE" with "neural-network-based operator learning" offers the potential to construct a neural operator that is both interpretable and highly expressive. PIKFNO is developed precisely based on this insight.

As shown in Figure1, PIKFNO introduces a structural modification to the trunk network of DeepONet: instead of learning arbitrary basis functions, the trunk network is constrained to radial basis function (RBF) form, with the RBF centers aligned with the boundary collocation points. This design makes the operator structure closer to the kernel expansion used in meshless methods, while avoiding the ill-conditioning issues caused by exterior source points in MFS. The physics-informed kernel function is denoted by $\psi(r)$, and its form depends on whether an analytical fundamental solution is available.

When the fundamental solution of the governing equation is difficult to derive but abundant measurement or numerical data are available, we learn the physics-informed kernel function $\psi(r)$ using a one-dimensional neural network:

$$\psi(r) = \mathrm{NN}(r;\theta) \tag{3}$$

where $\theta$ denotes the network parameters. After training, $\psi(r)$ can be regarded as a "numerical fundamental solution", it is nonsingular at the source point while retaining the implicit physical structure learned from data. Importantly, the learned kernel function can be used independently of PIKFNO, offering new possibilities for constructing other physics-based models.

When the fundamental solution $\varphi(r)$ is known, we do not use it directly. Instead, we define the physics-informed kernel function as

$$\psi(r) = k\varphi(r+\beta) + b \tag{4}$$

where $k$, $\beta$, and $b$ are learnable parameters, with $\beta > 0$ introduced to avoid the singularity at $r = 0$. The parameters $k$ and $b$ apply a linear transformation to the fundamental solution, which preserves its satisfaction of the governing equation while significantly improving the numerical range of the coefficients produced by the branch network. Directly using the unscaled fundamental solution often leads to coefficients $\omega_i$ spanning several orders of magnitude, resulting in unstable optimization or even non-convergence. By introducing learnable linear transformations, we maintain physical consistency while improving training stability and controllability.

In summary, PIKFNO explicitly embeds physics-informed kernel functions into the neural operator architecture, achieving a principled integration of physical priors and data-driven learning. Compared with traditional DeepONet, PIKFNO offers stronger interpretability, requires significantly fewer training samples for comparable accuracy, and exhibits faster convergence, providing a new pathway for constructing efficient, robust, and physically consistent neural operators.

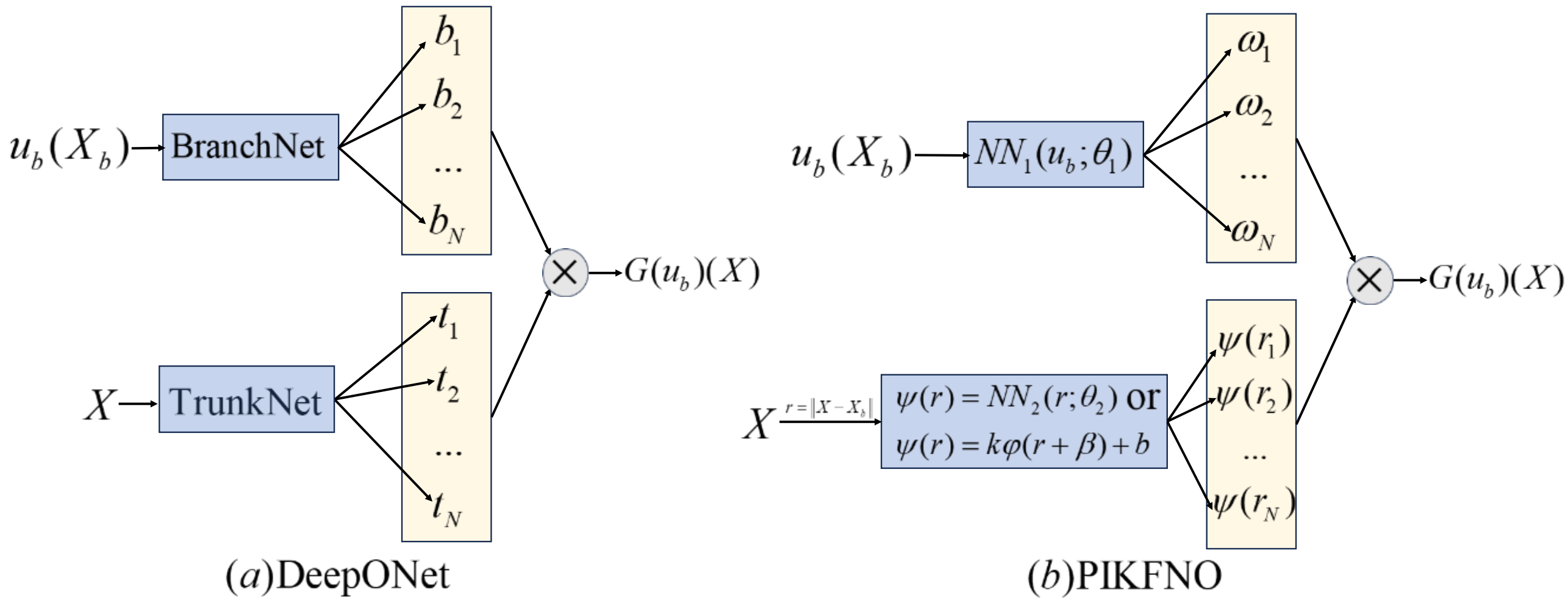

Figure 1 Network architectures of DeepONet and Physics-Informed Kernel Function Neural Operator

## 3 Numerical examples and discussions

In this section, we consider the numerical solution of the following two-dimensional Laplace equation:

$$\begin{cases} \Delta u(x,y)=0 & (x,y)\in\Omega \\ u(x,y)=u_b(x,y) & (x,y)\in\partial\Omega \end{cases} \tag{5}$$

where $\Delta$ denotes the Laplace operator, $\Omega=(0,1)\times(0,1)$ is the computational domain, $(x,y)$ represents the spatial coordinates, and $u_b$ is the prescribed Dirichlet boundary condition. Our objective is to learn the mapping from the boundary condition $u_b$ to the corresponding solution $u(x,y)$ using an operator-learning framework.

Following the setup commonly used in the DeepONet literature, we randomly generate 10,000 boundary conditions and compute the corresponding PDE solutions using a traditional numerical solver with a resolution of $40\times40$. Among these samples, 5,000 are used for training and the remaining 5,000 for testing. The input to the branch network consists of the discretized boundary values, where 40 points are sampled on each side of the boundary, yielding a total of 160 input features. The trunk network takes the spatial coordinates $(x,y)$ as input. The output of the neural operator is the numerical solution of the PDE under the given boundary condition.

Table 1 summarizes the network architectures of DeepONet and the two PIKFNO variants, together with their test-set error performance. All three models share an identical branch-network architecture, and their test errors are small and relatively close to each other. DeepONet achieves the highest prediction accuracy because its trunk network is capable of approximating functions in a much broader function space. For PIKFNO_v1, the trunk network learns within a radial basis function space, resulting in slightly lower accuracy than DeepONet but requiring significantly fewer learnable parameters. In contrast, the trunk network of PIKFNO_v2 is restricted to a linear combination of fundamental solutions, with only three learnable parameters $k$, $\beta$, and $b$. This yields the most limited expressive capacity among the three models and consequently leads to the largest test error.

Table 1 Network configurations and test set errors of DeepONet and the two PIKFNO variants

| | DeepONet | PIKFNO_v1 | PIKFNO_v2 |
|---|---|---|---|
| BranchNet | [160,160,160,160] | [160,160,160,160] | [160,160,160,160] |
| TrunkNet | [2,160,160,160] | $\psi(r) = \mathrm{NN}(r;\theta)$ ([1,160,160,1]) | $\psi(r) = k\varphi(r+\beta)+b$ |
| Test set error | $8.32\times10^{-4}$ | $8.90\times10^{-4}$ | $1.65\times10^{-3}$ |

Figure 2 presents the reference solution, the DeepONet prediction and its absolute error distribution, as well as the predictions and error distributions of the two PIKFNO models under a representative boundary condition. All three models produce solutions that closely match the reference solution, with uniformly small errors. The lower-left panel of Figure 2 further shows the evolution of the loss function during training for the three models. It can be observed that DeepONet and PIKFNO_v1 achieve similar final loss values, whereas PIKFNO_v2 attains the highest final loss. In terms of convergence speed, PIKFNO_v2 converges the fastest, followed by PIKFNO_v1, while DeepONet converges the slowest. This behavior can be explained as follows: PIKFNO_v2 employs the fundamental solution directly as the basis of its trunk network, effectively embedding part of the physical structure into the model, which leads to the fastest convergence. PIKFNO_v1 relaxes the physical prior to some extent, achieving relatively fast convergence while maintaining high accuracy. In contrast, DeepONet must search for suitable basis functions within a much broader function space, resulting in slower convergence but ultimately higher accuracy.

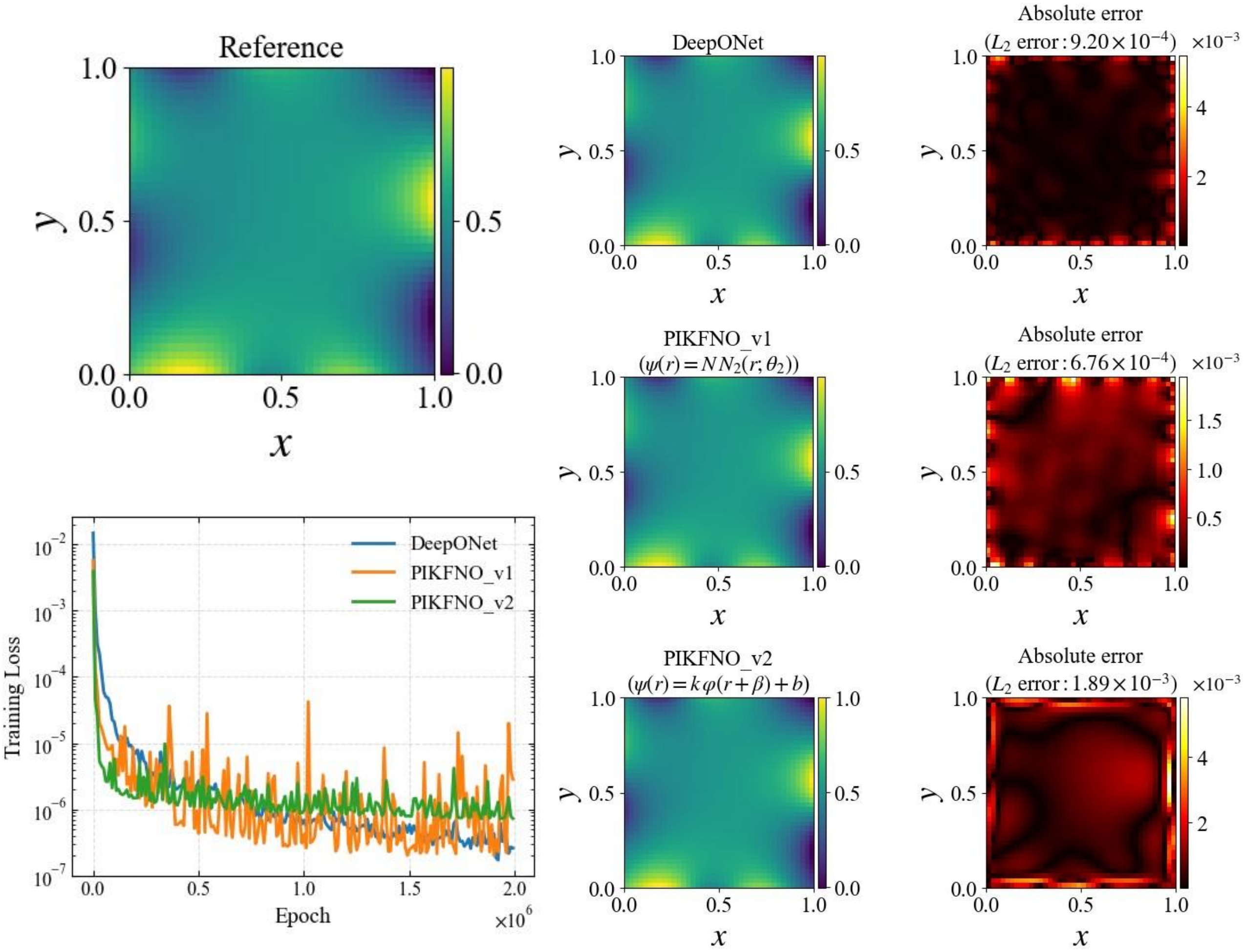

Figure 2 Reference solution under a specific boundary condition, DeepONet prediction and absolute error distribution, predictions and absolute error distributions of the two PIKFNO variants, and the training loss evolution of DeepONet and the two PIKFNO models

Figure 3 illustrates the kernel functions $\psi(r)$ obtained after training for the two PIKFNO models over the interval [0,2]. For PIKFNO_v1, the kernel function $\psi(r)$ is learned directly by a neural network. It can be observed that $\psi(r)$ exhibits a clear linear relationship with the fundamental solution $\varphi(r)$ of the Laplace equation. When a linear transformation of $\varphi(r)$ is plotted, it aligns remarkably well with the learned $\psi(r)$. However, unlike the true fundamental solution, $\psi(r)$ does not possess a singularity at $r=0$. In other words, PIKFNO_v1 learns a nonsingular “numerical fundamental solution” that is implicitly determined by the governing equation. For PIKFNO_v2, the kernel function $\psi(r)$ is defined as a linear transformation of the fundamental solution $\varphi(r)$, controlled by three learnable parameters $k$, $\beta$, and $b$. Although the two models yield kernel functions of different forms, a noteworthy observation is that their values at $r=0$ are nearly identical. From the perspective of the Singular Boundary Method, the value $\psi(0)$ is primarily determined by the location of the source point, which explains the strong consistency exhibited by both models near this point.

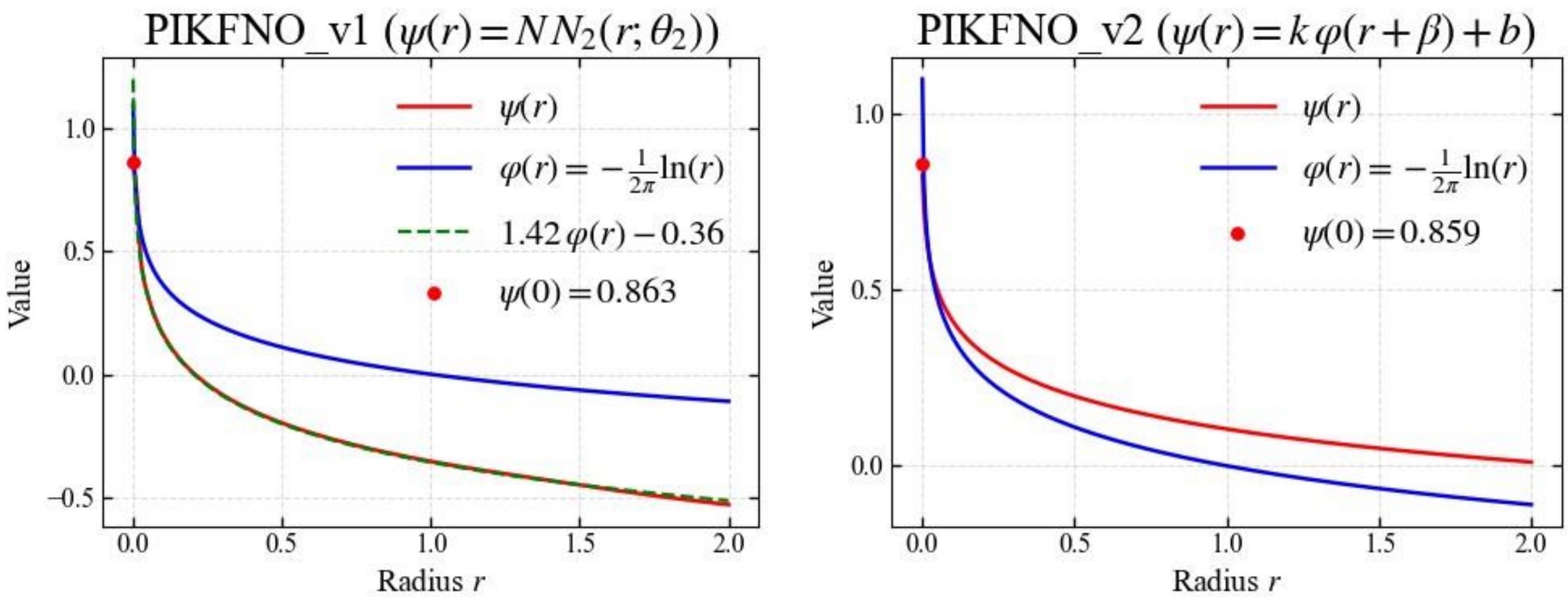

Figure 3 Visualization of the learned radial function $\psi(r)$ obtained from the two trained PIKFNO models over the interval [0,2], together with its comparison to the Laplace fundamental solution $\varphi(r)$.

For PIKFNO_v1, the learned kernel function $\psi(r)$ can be directly used as a fundamental solution. To verify this capability, we recompute a test case with an analytical solution given by $u(x,y)=x^3-3xy^2$. The boundary conditions are prescribed from the exact solution, and the computational domain is a disk centered at (0.5,0.5) with radius 0.5. The source-point configuration is identical to that used in the previous experiments. Figure 4 presents the exact solution, the numerical solution obtained by applying the Method of Fundamental Solutions (NN-based MFS) using the PIKFNO_v1 kernel $\psi(r)$ as the fundamental solution, and the corresponding relative error distribution. The NN-based MFS solution agrees extremely well with the exact solution, exhibiting only negligible errors. This demonstrates that the kernel function $\psi(r)$ learned by PIKFNO_v1 can indeed serve as a valid fundamental solution. In other words, PIKFNO_v1 provides a data-driven mechanism for discovering fundamental solutions, a capability that may have meaningful implications for the development of meshless methods.

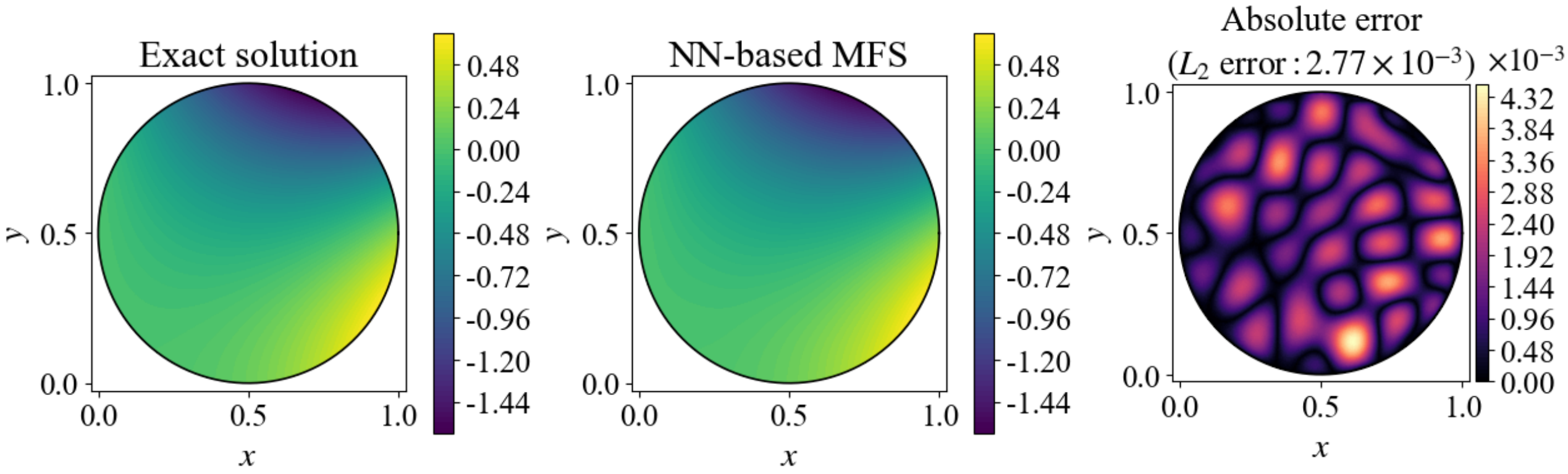

Figure 4 Exact solution, numerical solution obtained by Method of Fundamental Solutions using the learned $\psi(r)$ from the trained PIKFNO_v1 model as the fundamental solution (NN-based MFS), and the corresponding relative error distribution.

For PIKFNO_v2, we focus on its performance under limited training data. Table 2 reports the test errors of DeepONet and PIKFNO_v2 for different training-set sizes (with the test set fixed at 5000 samples). It can be observed that when the amount of training data is small, PIKFNO_v2 achieves significantly higher prediction accuracy than DeepONet. Many engineering problems—such as acoustic wave propagation—possess well-defined fundamental solution structures, and PIKFNO_v2 is able to exploit this prior knowledge to construct an efficient neural operator with substantially fewer data and lower computational cost.

Table 2 Test set errors of DeepONet and PIKFNO_v2 trained under varying training set sizes

| Training set size | 50 | 100 | 300 | 500 |
|---|---|---|---|---|
| DeepONet | $8.18\times10^{-2}$ | $4.41\times10^{-2}$ | $7.22\times10^{-3}$ | $3.57\times10^{-3}$ |
| PIKFNO_v2 | $1.58\times10^{-2}$ | $1.06\times10^{-2}$ | $6.55\times10^{-3}$ | $4.87\times10^{-3}$ |

## 4 Conclusions

In this study, we introduced the PIKFNO framework, which embeds physics-informed kernel functions directly into the neural operator architecture, enabling a principled integration of physical priors and data-driven learning. Compared with traditional neural operators, PIKFNO offers several notable advantages: (i) its operator structure aligns with the kernel expansions used in meshless methods, leading to enhanced interpretability; (ii) it maintains high predictive accuracy even with limited training data, with PIKFNO_v2 particularly benefiting from the explicit incorporation of fundamental-solution structures; and (iii) PIKFNO_v1 is capable of learning a nonsingular "numerical fundamental solution" from data, which can be directly used in the Method of Fundamental Solutions, revealing promising theoretical and practical implications. Numerical experiments confirm the effectiveness and robustness of the proposed framework. Future work will explore its application to more complex PDEs, nonlinear operators, and multiphysics problems, as well as further investigate the theoretical properties of data-driven fundamental solutions.

## Acknowledgments

The research was supported by the National Natural Science Foundation of China (12122205 and 12372196).